\documentclass{ecai} 

\usepackage{latexsym}
\usepackage{amssymb}
\usepackage{amsmath}
\usepackage{amsthm}
\usepackage{booktabs}
\usepackage{enumitem}
\usepackage{graphicx}
\usepackage{color}
\usepackage{listings}
\usepackage{xcolor}
\usepackage{multirow}

\usepackage{algorithm}
\usepackage{algpseudocode}
\usepackage{amsmath}
\usepackage{bm}

\newcommand{\BibTeX}{B\kern-.05em{\sc i\kern-.025em b}\kern-.08em\TeX}

\begin{document}


\begin{frontmatter}


\paperid{123} 


\title{AugMixCloak: A Defense against Membership Inference Attacks via Image Transformation}


\author{\fnms{Heqing}~\snm{Ren}}
\author{\fnms{Chao}~\snm{Feng}}
\author{\fnms{Alberto}~\snm{Huertas}}
\author{\fnms{Burkhard}~\snm{Stiller}} 

\address{University of Zurich}


\begin{abstract}

Traditional machine learning (ML) raises serious privacy concerns, while federated learning (FL) mitigates the risk of data leakage by keeping data on local devices. However, the training process of FL can still leak sensitive information, which adversaries may exploit to infer private data. One of the most prominent threats is the membership inference attack (MIA), where the adversary aims to determine whether a particular data record was part of the training set.

This paper addresses this problem through a two-stage defense called \textit{AugMixCloak}. The core idea is to apply data augmentation and principal component analysis (PCA)-based information fusion to query images, which are detected by perceptual hashing (pHash) as either identical to or highly similar to images in the training set. Experimental results show that \textit{AugMixCloak} successfully defends against both binary classifier-based MIA and metric-based MIA across five datasets and various decentralized FL (DFL) topologies. Compared with regularization-based defenses, \textit{AugMixCloak} demonstrates stronger protection. Compared with confidence score masking, \textit{AugMixCloak} exhibits better generalization.

\end{abstract}

\end{frontmatter}


\section{Introduction}

Traditional Machine Learning (ML) methods require all training data to be aggregated on a centralized entity, which inherently raises serious privacy and compliance concerns. To preserve privacy, Google proposed centralized federated learning (CFL)~\cite{bonawitz2017practical}. In this vanilla FL paradigm, each participant trains the model locally using its own local dataset, and shares model updates with the central aggregator. The aggregator then aggregates these model updates to update the global model that is sent back to the participants. This process is repeated until the global model converges or meets some stopping criteria. It helps protect user privacy by keeping data on local devices and improves the model's generalization ability through collaborative learning. However, CFL suffers from the single point of failure risk at the aggregator.

Decentralized federated learning (DFL) was proposed to address these issues~\cite{beltran2023decentralized}. DFL removes the need for a central aggregator; instead, participants directly exchange model updates. By leveraging the peer-to-peer communication mechanism, DFL provides stronger robustness against the single point of failure and improves overall processing capability.

Although CFL and DFL reduce the risk of data leakage compared to traditional ML, the training process still leaves behind residual data traces, which can lead to the leakage of sensitive information~\cite{bouacida2021vulnerabilities}. Attackers can exploit various information, such as gradient changes~\cite{nasr2019comprehensive} and prediction vectors, to infer private information. One of the most prominent types of such attacks is membership inference attack (MIA)~\cite{shokri2017membership}, where adversaries aim to determine whether a particular data record was included in the training set. Successfully executing MIA can reveal sensitive personal information.

To defend against MIAs, researchers have proposed approaches such as confidence score masking~\cite{jia2019memguard}, generative adversarial network (GAN)-based training~\cite{creswell2018generative}, differential privacy (DP)~\cite{ying2020privacy}, regularization~\cite{ying2020privacy}, and knowledge distillation~\cite{hinton2015distilling}, aiming to obscure the distinction between members and non-members. However, these defenses either incur high computational costs (e.g., GANs, knowledge distillation, DP) or provide limited protection despite being lightweight (e.g., confidence masking, regularization), as shown in Table~\ref{tab:defense_comparison}. Moreover, most of them are designed for traditional ML or CFL and lack adaptability to DFL, where the absence of a central coordinator and the heterogeneity of participants pose additional challenges. This highlights a pressing need for lightweight and DFL-compatible MIA defenses.

\begin{table}[ht]
\centering
\caption{Comparison of MIA Defense Approaches}
\label{tab:defense_comparison}
\begin{tabular}{llll}
\toprule
\textbf{Approach}               & \textbf{Target}      & \textbf{Effectiveness}      & \textbf{Cost} \\
\midrule
GAN                           & Dataset        & High         & High                         \\
Knowledge Distillation         & Model         & High                & High           \\
Differential Privacy           & Model          & High         & High                         \\
Confidence Score Masking        & Pred. vector    & Low      & Low                          \\
Regularization                  & Model        & Low          & Low                          \\
\textit{AugMixCloak} (This work)             & Query Image           & High                      & Low                          \\
\bottomrule
\end{tabular}
\end{table}
 
To address these limitations, this paper proposes a novel defense approach, named \textit{AugMixCloak}, which effectively mitigates MIAs without degrading the predictive performance for benign users and serves as a training-free reactive moving target defense (MTD). 

Different from existing approaches that operate on the entire dataset, model, or prediction vector, \textit{AugMixCloak} tackles the problem from a new perspective: query image transformation, which is simple, straightforward and very effective. It uses perceptual hashing (pHash) to monitor input similarity and detect potential attacks in real time. Upon detection, a two-stage defense is triggered: data augmentation is first applied to the input, followed by 
principal component analysis (PCA)-based information fusion~\cite{jolliffe2002principal} with representations derived from local training data. This dynamic transformation process obscures distinguishing features, making it significantly harder for attackers to infer membership. Unlike many existing defenses, \textit{AugMixCloak} operates purely at the test phase, introducing no modifications to the training process and incurring minimal computational overhead. The approach is deterministic, preventing exploitation through randomness and providing robust, reproducible protection. Extensive experiments across five datasets show that \textit{AugMixCloak} effectively mitigates both binary classifier-based and metric-based MIA variants, demonstrating strong performance under various DFL topologies while preserving model utility.


\section{Background and Related Work}
This section introduces the background of MIA and its threat to user privacy in FL environments. It further reviews the landscape of existing defense techniques, discussing their effectiveness and limitations, particularly in the context of decentralized federated settings.

\subsection{MIA on FL}

MIA was first implemented by Shokri et al. in 2017 to attack machine learning models. They demonstrated that an attacker can infer whether a specific data point was part of a model’s training dataset based on the model’s output \cite{shokri2017membership}.

The main reason why MIA can succeed is that the target model is overfitted \cite{chen2020gan}. Overfitting happens when the model not only learns the main features in the data but also remembers specific details and noise. As a result, the model performs significantly better on the training data than on the test data, because the noise and details are not generalizable \cite{hastie2009elements}.

There are two types of MIA:
\begin{itemize}
    \item Binary classifier based MIA: It consists of three steps: First, the attacker uses shadow training datasets to train multiple shadow models. Second, the attacker feeds the shadow training set and shadow test set into the trained shadow models to obtain their outputs, which correspond to members and non-members, respectively. Third, these two classes are then used to train a binary classifier.
    \item Metric based MIA: It calculates a metric for a given input record and directly compares it to a predetermined threshold to determine whether the input record is a member or non-member.
\end{itemize}

Attackers could also successfully conduct MIA by analyzing the local model updates of participants in FL. The first MIA against FL was introduced by Melis et al \cite{melis2019exploiting} in 2019. They found that the FL model shows specific gradient changes for training data and different gradient changes for test data, and this difference could be exploited to perform MIA. Since then, researchers have proposed more MIAs against FL \cite{zhang2020gan} \cite{gu2022cs}. It is worth noting that MIA aims to determine whether a data point belongs to the entire FL model’s dataset, rather than the dataset of a specific participant.

MIAs against FL can be categorized into two types. The first is a passive approach that follows the FL protocol to infer the membership of a data record \cite{truex2019demystifying}, meaning it does not alter the normal functioning of FL and instead passively collects and analyzes information to conduct the attack. The second is an active approach, in which the attacker tampers with the FL training process, thereby causing the model to leak more private data \cite{nasr2019comprehensive}.

\subsection{Existing Defense Approaches against MIA}

This section presents an overview of representative defense approaches against MIAs.

\textit{GAN} is made up of a generator and a discriminator. The generator produces synthetic data approximating the distribution of the original dataset, and the discriminator distinguishes between the original and generated data. Through continuous adversarial training, the similarity between the generated dataset and the original dataset gradually decreases \cite{hu2022defending}. Using generated dataset instead of the original dataset to train the target model, GANs effectively reduce the model’s ability to memorize individual samples from the original dataset, thereby enhancing defense against MIA.

\textit{Knowledge distillation} primarily reduces model overfitting. It first trains a large teacher model on the original dataset, where the model’s output for each sample is a probability distribution over classes, known as soft labels. The student model is then trained on these soft labels, gaining improved generalization ability and effectively defending against MIA. Some researchers also proposes new methods based on knowledge distillation, including repeated knowledge distillation \cite{mazzone2022repeated} and complementary knowledge distillation \cite{zheng2021resisting}.

\textit{Confidence score masking} refers to hiding the true confidence score of the target model by providing a modified confidence score, making it difficult for the attacker to determine whether the input record belongs to the training dataset based on the given confidence score. \cite{chen2023overconfidence} adopts a dual defense strategy: during the training phase, high-entropy soft labels are used to reduce the confidence of the target class; during the test phase, the original confidence score is replaced with a low-confidence score. \cite{jia2019memguard} introduces adversarial noise into the model’s confidence scores to protect privacy, thereby disrupting the attacker’s judgment.

\textit{DP} mitigates MIA by adding noise to the gradients or parameter updates of the model, making the influence of individual samples negligible, thereby altering the model's output. The noise can be either Laplace noise \cite{ying2020privacy} \cite{liu2020secure} or Gaussian noise \cite{chen2020differential} \cite{rahman2018membership} and \cite{liu2020secure}.

\textit{Regularization} defends against MIA by reducing the overfitting of the target model. Specific regularization methods include L2 regularization \cite{ying2020privacy}, dropout \cite{srivastava2014dropout}, and data augmentation \cite{kaya2021does}, all of which reduce overfitting by improving the model's generalization ability.

Although DFL provides stronger privacy preservation compared to traditional ML, existing defense approaches still do not provide strong protection in DFL settings. GAN-based methods, knowledge distillation, and DP require high computational overhead and are difficult to implement in decentralized environments due to the absence of a central coordinator. In contrast, techniques such as confidence score masking and regularization are much easier to deploy but provide limited protection in DFL. Moreover, the mechanisms behind confidence score masking, such as truncating, adding noise to, or smoothing the prediction vector, are relatively simple. As a result, adversaries may adapt to these perturbations through adversarial training, thereby improving their attack success rate.

Therefore, \textit{AugMixCloak} addresses the limitations of existing approaches by providing high effectiveness and low computational cost. It operates at the test phase by applying image transformations directly to the query image, providing a new perspective on defending against MIAs.

\section{Approach}


This paper proposes a novel, practical defense approach to protect DFL models against MIA. It performs data augmentation methods and PCA-based information fusion on the images existing in the training set before they are input into the target model, to prevent attackers from distinguishing between training set images and test set images. Besides, the approach selects specific data augmentation methods and a corresponding PCA-reconstructed image for each image based on its pHash value, ensuring that an attacker receives the same output when submitting the same image repeatedly.

\subsection{Defense Process}

The new defense approach consists of three steps.

\begin{figure}[H]
    \centering
    \includegraphics[width=0.95\linewidth]{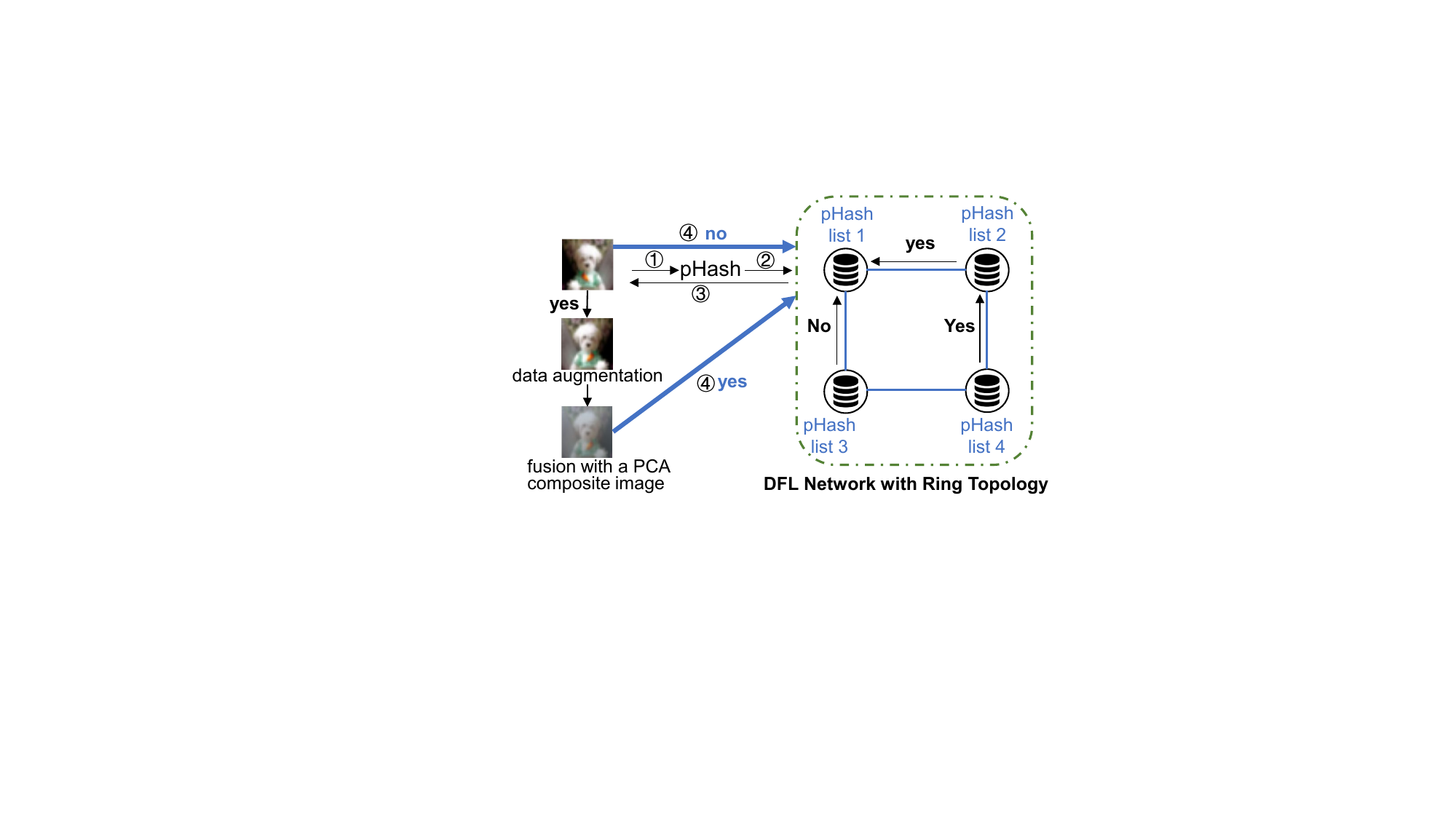}
    \caption{Methodology}    
    \label{fig:methodology}
    \vspace{-5pt}
\end{figure}

\textbf{Step 1: Calculate pHash value and then determine whether the query image is in the training set.} Each participant in the DFL topology computes the pHash values of all images in its local dataset in advance to obtain a local pHash list. As shown in Figure~\ref{fig:methodology}, pHash list 1-4 are the local pHash lists corresponding to participants 1, 2, 3, and 4 in a DFL network with ring topology, respectively. 

When an attacker queries an image at participant 1, the pHash of the image is first calculated, as shown in phase 1 of Figure~\ref{fig:methodology}. Then, participant 1 performs a binary search in its local pHash list to check for the presence of this pHash. If it is found, it returns \textit{yes}. If not, participant 1 queries its neighbors, i.e., participant 2 and participant 3, to see whether they have the pHash in their local pHash lists. Participant 2 also queries its neighbor, participant 4. If any participant’s neighbor returns \textit{yes}, then participant will also return \textit{yes}. In other words, as long as the pHash of the queried image exists in any participant’s local pHash list, the final result will be \textit{yes}; otherwise, it will be \textit{no}, as shown in phase 3. If the result is \textit{no}, the original query image is directly input to the target model. If the result is \textit{yes}, the original image is first processed by data augmentation and fusion with a PCA-reconstructed image, and then the processed image is input to the target model, as shown in phase 4.

\begin{figure}[htbp]
    \centering
    \begin{minipage}[t]{0.48\linewidth}
        \centering
        \includegraphics[width=\linewidth]{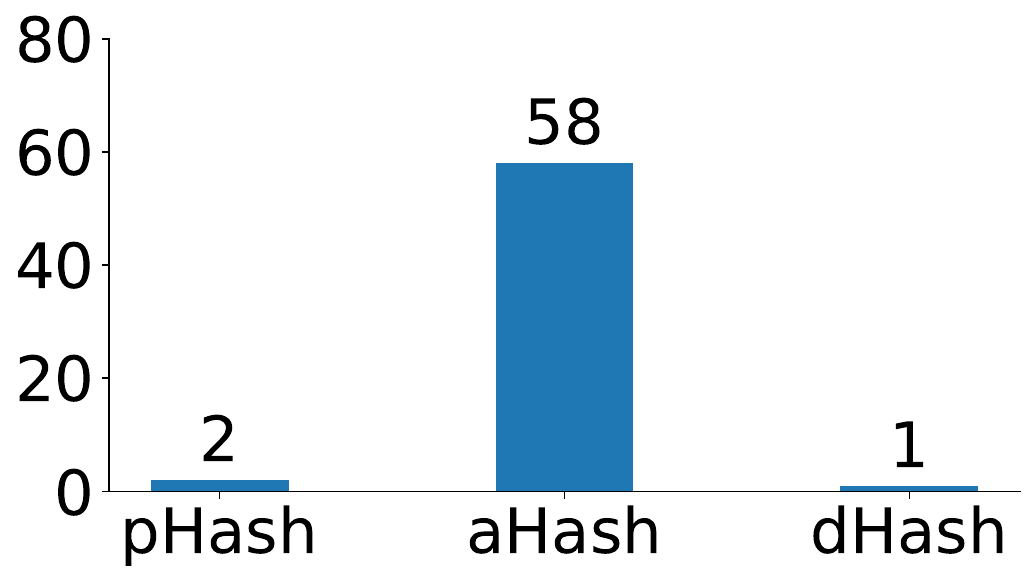}
        \vspace{-20pt}
        \caption{Duplicate Hashes in Training Set}
        \label{fig:duplicate_hashes}
    \end{minipage}
    \hfill
    \begin{minipage}[t]{0.48\linewidth}
        \centering
        \includegraphics[width=\linewidth]{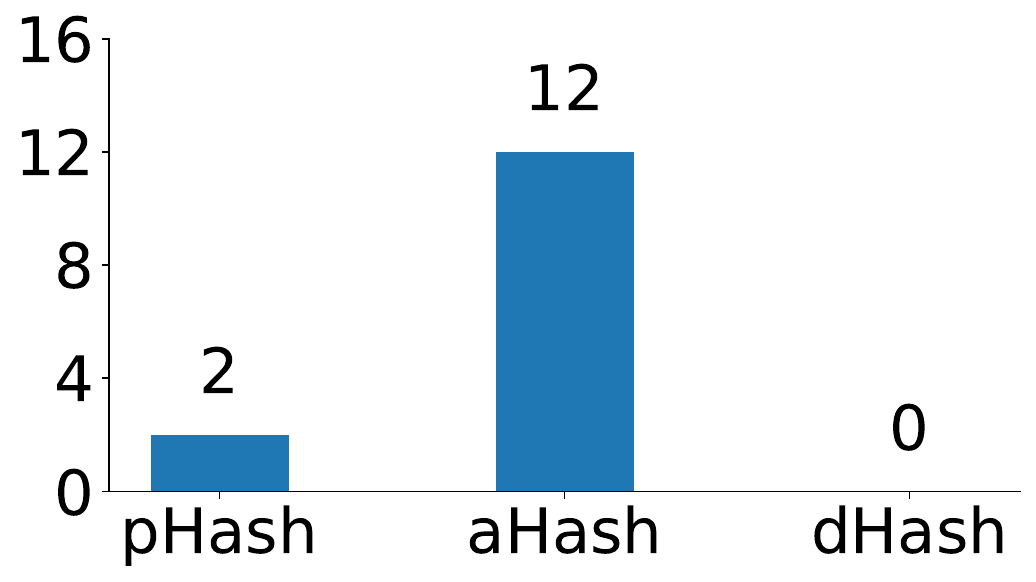}
        \vspace{-20pt}
        \caption{Test Hash Occurrence in Training Set}
        \label{fig:test_hash_frequency}
    \end{minipage}
\end{figure}

There are two primary reasons for choosing pHash over average hashing (aHash) and difference hashing (dHash). First, pHash assigns identical hash values to images with slight differences. Therefore, query images that are highly similar to a training set image are also likely to be identified as part of the training set. Second, the number of duplicate pHash values in the training dataset and the frequency with which a test sample's pHash appears in the training set's pHash list are relatively low. Two bar charts, Figure~\ref{fig:duplicate_hashes} and~\ref{fig:test_hash_frequency}, illustrate the comparison results when using the CIFAR-10 dataset.

\textbf{Step 2: Perform specific data augmentation methods.} As shown in Algorithm~\ref{alg:aug}, this approach employs 12 image augmentation methods, such as horizontal image flipping, image rotation, image sharpening, \emph{etc}. The approach selects specific data augmentation methods for each query image, based on its pHash value, in which \textit{aug\_key} is obtained by the following Equation \ref{eq:data_aug_key}. After performing selected data augmentation methods on the query image, a data-augmented image is generated, and then proceed to the next step.

For example, if the query image has a pHash value of 24648, then the corresponding \textit{aug\_key} is 0. If \textit{aug\_num = 2}, the approach will select the first and second data augmentation methods, \textit{HorizontalFlip} and \textit{Rotation}, to perform on the original query image.

\begin{equation}\label{eq:data_aug_key}
\textit{aug\_key} = \textit{phash\_decimal} \bmod 12
\end{equation}

\begin{algorithm}[htbp]
\caption{Apply Data Augmentation}
\label{alg:aug}
\begin{algorithmic}[1]
\State \textbf{Input:} Image $image$, Integer $aug\_num$, Integer $aug\_key$
\State \textbf{Output:} Augmented Image
\State \parbox[t]{\dimexpr\linewidth-\algorithmicindent}{
$augs \gets [$\textit{HorizontalFlip, Rotation, AffineTranslate, Affine, CenterCrop, Perspective, Equalize, CenterCrop, GaussianBlur, Grayscale, Sharpening, Posterize}$]$}
\State $total \gets length(augs)$
\State $selected \gets$ empty list
\For{$i = 0$ to $aug\_num - 1$}
  \State $idx \gets (aug\_key + i) \bmod total$
  \State $selected \gets selected \cup \{augs[idx]\}$
\EndFor
\State $selected\_augs \gets Compose(selected)$
\State $augmented\_image \gets selected\_augs(image)$
\State \Return $augmented\_image$
\end{algorithmic}
\end{algorithm}

\textbf{Step 3: Perform fusion with a specific PCA-reconstructed image.} The approach generates a PCA-reconstructed image using the first principal component for each class of images in the database. For example, there are 100 classes in the CIFAR-100 dataset, so 100 PCA-reconstructed images are generated for it. To build a PCA-reconstructed image for each class, the approach first collects all training images belonging to the same class and flatten them into vectors. Specifically, given \( n \) images of class \( c \), each of size \( H \times W \times C \), the approach reshapes them into row vectors and stacks them to form the data matrix:

\begin{equation}
X = \begin{bmatrix}
\text{vec}(I_1) \\
\text{vec}(I_2) \\
\vdots \\
\text{vec}(I_n)
\end{bmatrix} \in \mathbb{R}^{n \times d}, \quad d = H \times W \times C.
\label{eq:pca_matrix}
\end{equation}

This approach then performs PCA on \( X \), and extracts the first principal component vector \( \mathbf{v}_1 \in \mathbb{R}^{d} \), which captures the direction of maximum variance in the class-specific image distribution:
\begin{equation}
\mathbf{v}_1 = \text{FirstPrincipalComponent}(X).
\label{eq:first_pc}
\end{equation}

To build a PCA-reconstructed image from this component, this approach scales and shifts \( \mathbf{v}_1 \) using the standard deviation \( \sigma_X \) and mean \( \mu_X \) of the data matrix:
\begin{equation}
\mathbf{m} = \mathbf{v}_1 \cdot \sigma_X + \mu_X.
\label{eq:reconstruct}
\end{equation}

Finally, the resulting vector \( \mathbf{m} \) is reshaped back to the original image shape to obtain the PCA-reconstructed image, which retains the dominant structural features of its corresponding class:
\begin{equation}
I_{\text{pca}} = \text{reshape}(\mathbf{m}, H, W, C).
\label{eq:reshape}
\end{equation}

 The approach selects a specific PCA-reconstructed image for each query image based on its pHash value. \textit{PCA\_key} is computed as shown in Equation~\ref{eq:pca_key}, where $n_{cls}$ represents the number of classes in the dataset, then this approach uses it to select a specific PCA-reconstructed image. After that, each pixel value of the data-augmented image is linearly combined with the corresponding pixel value of the selected PCA-reconstructed image with a fusion weight, to obtain the final image as shown in Equation~\ref{eq:fusion}, in which $ I_{\text{fused}}(i, j) $ denotes the pixel value at position $ (i, j) $ in the image, $ I_{\text{aug}}(i, j) $ is from the augmented image, and $ I_{\text{pca}}(i, j) $ is from the selected PCA-reconstructed image.

\begin{equation}\label{eq:pca_key}
\textit{PCA\_key} = \textit{phash\_decimal} \bmod n_{cls}
\end{equation}
\begin{equation}\label{eq:fusion}
I_{\text{fused}}(i, j) = \alpha \cdot I_{\text{aug}}(i, j) + (1 - \alpha) \cdot I_{\text{pca}}(i, j)
\end{equation}

\begin{figure}[h]
    \centering
    \includegraphics[width=0.95\linewidth]{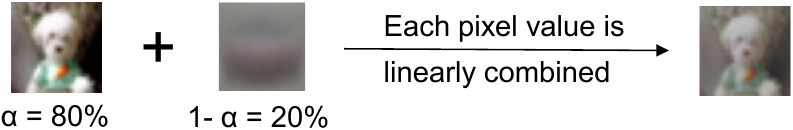}
    \caption{Performing Fusion with a PCA-Reconstructed Image}    
    \label{fig:PCA_composite}
\end{figure}

Figure~\ref{fig:PCA_composite} illustrates the fusion process between a data-augmented image and a PCA-reconstructed image. The image on the left is a horizontally flipped dog image from the CIFAR-10 dataset. The center image is a PCA-reconstructed image generated from the automobile class in the CIFAR-10 dataset, which clearly reveals distinctive features of automobiles. These two images are linearly combined with $\alpha = 80\%$, indicating that 80\% of the left image and 20\% of the center image are used, resulting in the processed final image shown on the right. Despite the fusion, it can still be recognized as a dog image, which implies that our defense approach effectively preserves the target model’s prediction accuracy on training images. As for the test images, they do not go through Step 2 or Step 3 at all, so their prediction accuracy remains completely unaffected.

\subsection{Automatic Determination of Defense Intensity}\label{sec:automatic}

After observing and validating multiple sets of experimental data, it is found that the F1-score of both binary classifier-based MIA and metric-based MIA increases with the rise of the PCA-based information fusion weight ($\alpha$), and decreases as the data augmentation intensity increases. The observed experimental results are illustrated in Section~\ref{sec:impact}.

Afterwards, this approach integrates a script to automatically find the appropriate defense intensity, which facilitates other researchers in quickly obtaining the defense intensity on different datasets.

The core idea of the script is to iterate over different values of $\alpha$, and for each $\alpha$, identify the predefined defense parameter combination $(\textit{num}, \textit{weights})$ that yields F1-scores within the range of [0.35, 0.65] and has the smallest deviation from 0.5, denoted as $best\_config$. It then refines the parameter combination based on $best\_config$ to bring the F1-scores even closer to 0.5, and finally selects the configuration with the overall smallest deviation.

\section{Evaluation Setup}

\subsection{Models and Datasets}

This paper evaluats three types of DFL topologies: fully connected, ring, and star, as detailed below, and uses 10 participants for each topology. 

\begin{itemize} 
    \item \textbf{Fully connected:} Each participant is directly connected to all other participants. In each round of communication, a participant exchanges model parameters with every other participant. 
    \item \textbf{Ring:} Each participant is connected to two neighbors, forming a closed loop. In each round, a participant only communicates with its two neighbors. 
    \item \textbf{Star:} One center participant is connected to all leaf participants, and the leaf participants are not connected to each other. The center participant exchanges model parameters with all other participants, while each leaf participant only communicates with the center participant. 
\end{itemize}

In the fully connected and ring topologies, all participants are structurally equal, so this paper sets the attacker at the first participant as a representative. In the star topology, the center participant has more information than the leaf participants, so this paper places the attacker at the center and also at the first leaf participant.

To calculate the global model, this paper uses average aggregation after selecting the participants.

This paper utilizes five widely used image classification datasets and deploys various classification models under the DFL framework to achieve effective predictive performance. The MNIST dataset is not included because MIA attacks are difficult to succeed on MNIST, making the use of \textit{AugMixCloak} meaningless.

\textbf{CIFAR-10.} The CIFAR-10 dataset consists of 60,000 color images of size 32×32 across 10 classes, with 50,000 images used for training and 10,000 for testing. This paper uses a lightweight CNN model to classify the dataset. In the fully connected topology, training runs for 15 rounds with 10 local epochs per round. In the star and ring topologies, it runs for 50 rounds with 15 local epochs per round because their information propagation is less efficient than that of the fully connected topology.

\textbf{CIFAR-100.} The CIFAR-100 dataset contains the same images as CIFAR-10, but they are categorized into 100 classes instead of 10. Due to the larger number of classes and increased classification difficulty, this paper adopts VGG16 as the classification model. In the three DFL topologies, the total number of training rounds is 15, with 10 local training epochs per round. In the fully connected topology, training runs for 15 rounds with 10 local epochs per round, and in the star and ring topologies, it runs for 25 rounds with 15 local epochs per round.

\textbf{Fashion-MNIST.} The Fashion-MNIST dataset consists of 70,000 grayscale images of size 28×28 across 10 classes, with 60,000 images used for training and 10,000 for testing. This paper uses a lightweight CNN model to classify this dataset. In the fully connected topology, training runs for 20 rounds with 5 local epochs per round, and in the star and ring topologies, it runs for 50 rounds with 15 local epochs per round.

\textbf{Tiny-ImageNet.}  The Tiny-ImageNet dataset consists of 110,000 color images of size 64×64 across 200 classes, with 100,000 images used for training and 10,000 for testing. This paper employs ResNet50 to perform the classification task. In the fully connected topology, training runs for 15 rounds with 10 local epochs per round, and in the star and ring topologies, it runs for 30 rounds with 15 local epochs per round.

\textbf{ImageNet-10.} The ImageNet-10 dataset consists of 13,500 color images of various size across 10 classes, with 13,000 images used for training and 500 for testing. The classification model is based on the ResNet18 architecture. In the fully connected topology, training runs for 10 rounds with 5 local epochs per round, and in the star and ring topologies, it runs for 20 rounds with 10 local epochs per round.

\subsection{Data Preprocessing}

This paper uses Independent and Identically Distributed (IID) data partitions. To simulate the IID data distribution, all datasets are shuffled by class and then evenly distributed to each participant.

The image sizes within CIFAR-10, CIFAR-100, Fashion-MNIST, or Tiny-ImageNet datasets are identical. Therefore, the resizing operation is only required for the ImageNet-10 dataset.

Next, all datasets need to undergo the transformation process, including two steps: pixel value normalization to the range [0, 1] and standardization. After that, the processed images can be fed into the target model.

\subsection{MIA}

Two types of MIA are employed in this paper: binary classifier-based MIA and metric-based MIA. The metric-based MIA includes prediction correctness-based MIA, prediction entropy-based MIA, and prediction modified entropy-based MIA. The prediction modified entropy-based MIA modifies the prediction vector in the formula of the prediction entropy-based MIA to improve attack performance.

This paper evaluates the performance of MIA attacks using the F1-score. The further the values deviate from 0.5, the better the attack performance. After applying defenses, the closer the values are to 0.5, the more effective the defense is.

By comparing the values of F1-score with and without defense, this paper evaluates the effectiveness of \textit{AugMixCloak}. In addition, by comparing these metrics between \textit{AugMixCloak} and existing approaches, the effectiveness of \textit{AugMixCloak} can be further demonstrated.





\subsection{Defense}

In addition to experimenting with \textit{AugMixCloak}, this paper also evaluates two existing defense approaches: regularization and confidence score masking.

For regularization, L2 regularization is specifically employed. The main idea of L2 regularization is to add an additional term to the loss function that penalizes large model parameters. This encourages the model to learn simpler patterns, reduces overfitting, and ultimately smooths the output confidence scores.

For confidence score masking, the clipping method is adopted. The core idea of clipping is to limit the model’s maximum Softmax confidence score to a predefined threshold, thereby reducing the amount of information available to the attacker.

\section{Evaluation Results}

This section presents a comprehensive evaluation of \textit{AugMixCloak}, including the selection of defense parameters, its effectiveness across different topologies and datasets, the impact of data augmentation intensity and fusion weight, and comparisons with existing defense approaches such as regularization and confidence score masking.

\subsection{Defense Parameters}

\textit{AugMixCloak} integrates a script to automatically find the appropriate defense intensity as described in section~\ref{sec:automatic}.

The defense intensity for each dataset is shown in the Table~\ref{tab:defense_params}. Topology indicates the DFL topology, where star (center) represents the center participant in the star topology, and star (leaf) represents the leaf participant in the star topology. Intensity shows the number of data augmentation methods performed on the query image. For example, \textit{n = [1, 2]} and \textit{w = [0.9, 0.1]} shows a defense parameter combination (num, weights), indicating that there is a 90\% probability of selecting one data augmentation method and a 10\% probability of selecting two.

The number of data augmentation methods applied to each specific image is not chosen randomly but is also determined by the pHash value of the query image, ensuring the reproducibility of the results. $\alpha$ represents the PCA composite image fusion weight.

For certain rows in the table, the values of Data Augmentation Intensity and $\alpha$ are marked as -, indicating that for the corresponding dataset and topology, MIA could not work even without applying \textit{AugMixCloak}.

\begin{table}[!ht]
\centering
\caption{Defense Parameters Across Datasets and Topologies}
\label{tab:defense_params}
\begin{tabular}{cclc}
\toprule
\textbf{Dataset} & \textbf{Topology} & \textbf{Intensity} & \textbf{$\bm{\alpha}$ } \\
\midrule
\multirow{4}{*}{CIFAR-10}      
    & fully  & n = [0, 1], w = [0.7, 0.3] & 0.6 \\
    & ring   & -                                 & -   \\
    & star (center) & n = [1, 2], w = [0.5, 0.5] & 0.9 \\
    & star (leaf)   & n = [0, 1], w = [0, 1]     & 0.8 \\
\midrule
\multirow{4}{*}{CIFAR-100}     
    & fully  & n = [0, 1], w = [0, 1]     & 0.9 \\
    & ring   & n = [0, 1], w = [0.5, 0.5]     & 0.5 \\
    & star (center) & -                            & -   \\
    & star (leaf)   & -                            & -   \\
\midrule
\multirow{4}{*}{Fashion-MNIST} 
    & fully  & n = [0, 1], w = [0, 1] & 0.9 \\
    & ring   & n = [0, 1], w = [0, 1]     & 0.9 \\
    & star (center) & n = [0, 1], w = [0.3, 0.7] & 0.9 \\
    & star (leaf)   & n = [0, 1], w = [0, 1]     & 0.9 \\
\midrule
\multirow{4}{*}{Tiny-ImageNet} 
    & fully  & n = [0, 1], w = [0, 1]     & 0.9 \\
    & ring   & n = [0, 1], w = [0.3, 0.7] & 0.9 \\
    & star (center) & n = [0, 1], w = [0.4, 0.6] & 0.7 \\
    & star (leaf)   & n = [0, 1], w = [0.2, 0.8] & 0.9 \\
\midrule
\multirow{4}{*}{ImageNet-10}   
    & fully  & n = [2, 3], w = [0, 1] & 0.8 \\
    & ring   & n = [1, 2], w = [0.9, 0.1] & 0.9 \\
    & star (center) & n = [2, 3], w = [0.9, 0.1] & 0.9 \\
    & star (leaf) & n = [1, 2], w = [0.4, 0.6] & 0.9 \\
\bottomrule
\end{tabular}
\end{table}

\subsection{Evaluation Results of \textit{AugMixCloak}}

Table~\ref{tab:fully_connected} shows the result of fully connected topology. DSet represents the dataset. D1, D2, D3, D4, and D5 correspond to CIFAR-10, CIFAR-100, Fashion-MNIST, Tiny-ImageNet, and ImageNet-10, respectively. 
Def stands for defense: No indicates that \textit{AugMixCloak} is not applied, while Yes indicates that \textit{AugMixCloak} is applied. Acc1 and Acc2 denote the prediction accuracy on training and test samples, respectively. 
Binary is the F1-score of the binary classifier-based MIA. 
M1, M2, and M3 are the F1-scores of the prediction correctness-based, prediction entropy-based, and prediction modified entropy-based MIA, respectively.

By comparing the "Yes" rows with the "No" rows in Table ~\ref{tab:fully_connected}, i.e., the results after applying \textit{AugMixCloak} versus before applying it, the following findings are observed, which consistently hold across all five datasets:

First, after applying \textit{AugMixCloak}, Acc1 shows a significant drop. This is because the classification model's predictive capability decreases due to data augmentation and fusion with a PCA composite image. Moreover, the gap between Acc1 and Acc2 becomes small after the defense is applied, which further confuses the attackers.

Second, after applying \textit{AugMixCloak}, Acc2 changes very little. This indicates that \textit{AugMixCloak} does not affect the prediction accuracy for images submitted by benign users. The slight variation is due to a very small number of test images whose pHash values happen to appear in the pHash list of the training dataset.

Third, after applying \textit{AugMixCloak}, both the F1-score of the binary classifier-based MIA and that of the three metric-based MIA across all datasets fall within the range of 0.4 to 0.6. This suggests that neither type of MIA can distinguish between members and non-members, meaning \textit{AugMixCloak} successfully defends against both types of MIA.

\begin{table}[H]
\centering
\caption{Fully Connected Topology Results with and without Defense}
\label{tab:fully_connected}
\begin{tabular}{cccccccc}
\toprule
\textbf{DSet} & \textbf{Def} & \textbf{Acc1} & \textbf{Acc2} & \textbf{Binary} & \textbf{M1} & \textbf{M2} & \textbf{M3} \\
\midrule

\multirow{2}{*}{D1}      
& No & 93.5\% & 64.9\% & 0.663 & 0.703 & 0.642 & 0.703 \\
& Yes & 73.1\% & 64.9\% & 0.528 & 0.522 & 0.460 & 0.522 \\
\hline

\multirow{2}{*}{D2}     
& No & 99.5\% & 60.8\% & 0.756 & 0.770 & 0.732 & 0.771 \\
& Yes & 68.4\% & 60.8\% & 0.523 & 0.502 & 0.478 & 0.502 \\
\hline

\multirow{2}{*}{D3} 
& No & 97.9\% & 90.7\% & 0.603 & 0.668 & 0.590 & 0.666 \\
& Yes & 81.6\% & 90.6\% & 0.517 & 0.526 & 0.433 & 0.522 \\
\hline

\multirow{2}{*}{D4} 
& No & 97.8\% & 56.4\% & 0.744 & 0.772 & 0.734 & 0.771 \\
& Yes & 66.9\% & 56.4\% & 0.509 & 0.520 & 0.489 & 0.521 \\
\hline

\multirow{2}{*}{D5}   
& No & 98.6\% & 92.4\% & 0.647 & 0.641 & 0.627 & 0.641 \\
& Yes & 88.7\% & 92.4\% & 0.571 & 0.451 & 0.436 & 0.453 \\
\bottomrule
\end{tabular}
\end{table}

Tables~\ref{tab:ring}, \ref{tab:star_center}, and \ref{tab:star_leaf} present the evaluation results under the ring topology, and for the center and leaf participants in the star topology, respectively. In each case, the assigned defense intensity is determined by the script, as shown in Table~\ref{tab:defense_params}. Notably, MIA fails to succeed on CIFAR-10 in the ring topology and on CIFAR-100 for both the center and leaf participants in the star topology, even without applying \textit{AugMixCloak}. Therefore, the defense is not applied for these specific combinations.

\begin{table}[H]
\centering
\caption{Ring Topology Results with and without Defense}
\label{tab:ring}
\begin{tabular}{cccccccc}
\toprule
\textbf{DSet} & \textbf{Def} & \textbf{Acc1} & \textbf{Acc2} & \textbf{Binary} & \textbf{M1} & \textbf{M2} & \textbf{M3} \\
\midrule

\multirow{2}{*}{D1}      
& No & 67.7\% & 54.9\% & 0.544 & 0.473 & 0.479 & 0.472 \\
& Yes & - & - & - & - & - & - \\
\hline

\multirow{2}{*}{D2}     
& No & 78.7\% & 51.7\% & 0.615 & 0.608 & 0.583 & 0.610 \\
& Yes & 57.7\% & 51.8\% & 0.481 & 0.430 & 0.439 & 0.431 \\
\hline

\multirow{2}{*}{D3} 
& No & 94.0\% & 88.5\% & 0.623 & 0.641 & 0.619 & 0.639 \\
& Yes & 77.7\% & 88.4\% & 0.528 & 0.484 & 0.467 & 0.479 \\
\hline

\multirow{2}{*}{D4} 
& No & 90.2\% & 52.3\% & 0.709 & 0.631 & 0.616 & 0.631 \\
& Yes & 71.4\% & 52.3\% & 0.578 & 0.443 & 0.434 & 0.443 \\
\hline

\multirow{2}{*}{D5}   
& No & 94.5\% & 88.0\% & 0.585 & 0.596 & 0.584 & 0.598 \\
& Yes & 89.7\% & 88.0\% & 0.517 & 0.495 & 0.478 & 0.499 \\
\bottomrule
\end{tabular}
\end{table}

\begin{table}[H]
\centering
\caption{Results of the Center Participant in the Star Topology with and without Defense}
\label{tab:star_center}
\begin{tabular}{cccccccc}
\toprule
\textbf{DSet} & \textbf{Def} & \textbf{Acc1} & \textbf{Acc2} & \textbf{Binary} & \textbf{M1} & \textbf{M2} & \textbf{M3} \\
\midrule

\multirow{2}{*}{D1}      
& No & 96.1\% & 63.7\% & 0.662 & 0.694 & 0.599 & 0.693 \\
& Yes & 60.8\% & 63.7\% & 0.586 & 0.431 & 0.404 & 0.431 \\
\hline

\multirow{2}{*}{D2}     
& No & 86.1\% & 54.0\% & 0.496 & 0.446 & 0.401 & 0.444 \\
& Yes & - & - & - & - & - & - \\
\hline

\multirow{2}{*}{D3} 
& No & 97.9\% & 89.7\% & 0.571 & 0.656 & 0.587 & 0.656 \\
& Yes & 85.7\% & 89.6\% & 0.510 & 0.543 & 0.444 & 0.543 \\
\hline

\multirow{2}{*}{D4} 
& No & 97.6\% & 52.9\% & 0.745 & 0.697 & 0.676 & 0.696 \\
& Yes & 72.6\% & 52.9\% & 0.521 & 0.442 & 0.429 & 0.441 \\
\hline

\multirow{2}{*}{D5}   
& No & 96.4\% & 87.6\% & 0.607 & 0.635 & 0.610 & 0.638 \\
& Yes & 86.6\% & 87.6\% & 0.527 & 0.518 & 0.476 & 0.520 \\
\bottomrule
\end{tabular}
\end{table}

\begin{table}[H]
\centering
\caption{Results of the Leaf Participant in the Star Topology with and without Defense}
\label{tab:star_leaf}
\begin{tabular}{cccccccc}
\toprule
\textbf{DSet} & \textbf{Def} & \textbf{Acc1} & \textbf{Acc2} & \textbf{Binary} & \textbf{M1} & \textbf{M2} & \textbf{M3} \\
\midrule

\multirow{2}{*}{D1}      
& No & 91.1\% & 63.3\% & 0.638 & 0.681 & 0.604 & 0.681 \\
& Yes & 65.9\% & 63.3\% & 0.589 & 0.468 & 0.414 & 0.467 \\
\hline

\multirow{2}{*}{D2}     
& No & 70.6\% & 46.4\% & 0.495 & 0.451 & 0.404 & 0.450 \\
& Yes & - & - & - & - & - & - \\
\hline

\multirow{2}{*}{D3} 
& No & 96.8\% & 90.1\% & 0.570 & 0.664 & 0.623 & 0.665 \\
& Yes & 80.1\% & 89.9\% & 0.518 & 0.529 & 0.466 & 0.529 \\
\hline

\multirow{2}{*}{D4} 
& No & 94.7\% & 52.2\% & 0.711 & 0.668 & 0.649 & 0.665 \\
& Yes & 70.6\% & 52.2\% & 0.529 & 0.457 & 0.447 & 0.455 \\
\hline

\multirow{2}{*}{D5}   
& No & 93.6\% & 80.8\% & 0.595 & 0.619 & 0.586 & 0.619 \\
& Yes & 86.7\% & 80.8\% & 0.530 & 0.515 & 0.468 & 0.519 \\
\bottomrule
\end{tabular}
\end{table}

\subsection{Impact of Data Augmentation Intensity and $\alpha$}\label{sec:impact}

This paper investigates the impact of data augmentation intensity and PCA composite image fusion weight $\alpha$ on the prediction accuracy and F1-scores of both binary classifier-based MIA and the three metric-based MIAs.

\begin{figure}[htbp]
    \centering
    \begin{minipage}[t]{0.48\linewidth}
        \centering
        \includegraphics[width=\linewidth]{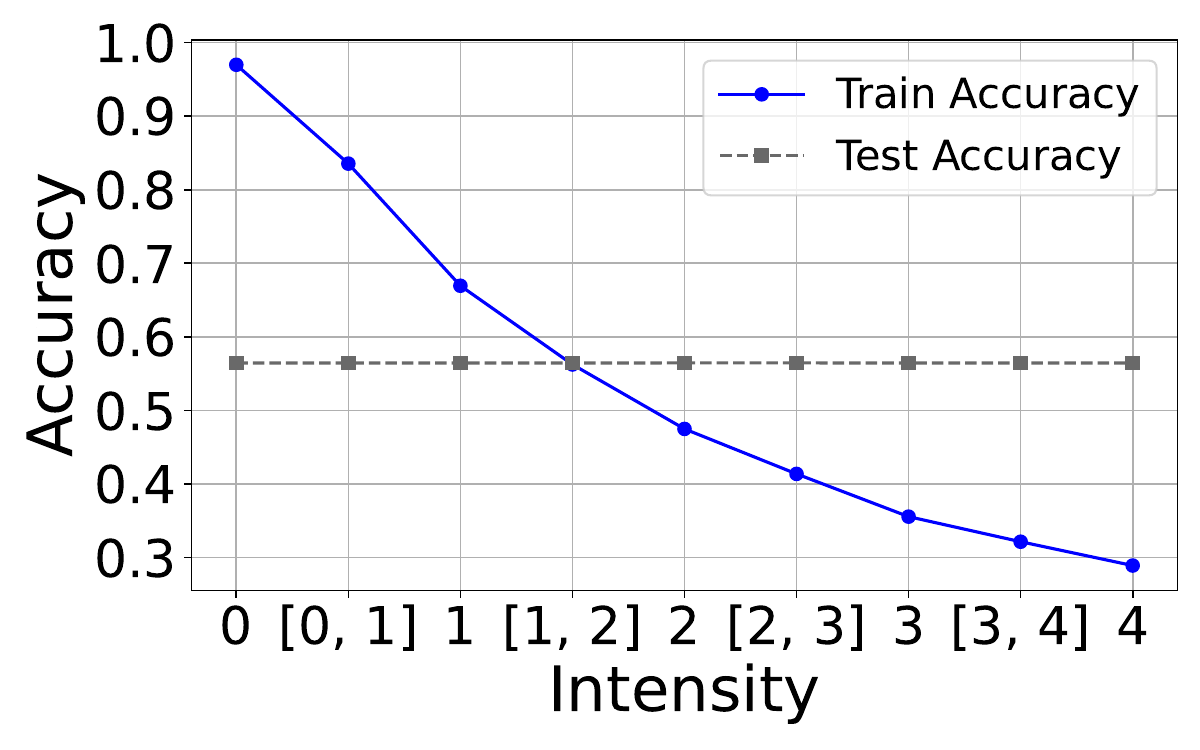}
        \vspace{-20pt}
        \caption{Accuracy vs Intensity}
        \label{fig:line1}
    \end{minipage}
    \hfill
    \begin{minipage}[t]{0.48\linewidth}
        \centering
        \includegraphics[width=\linewidth]{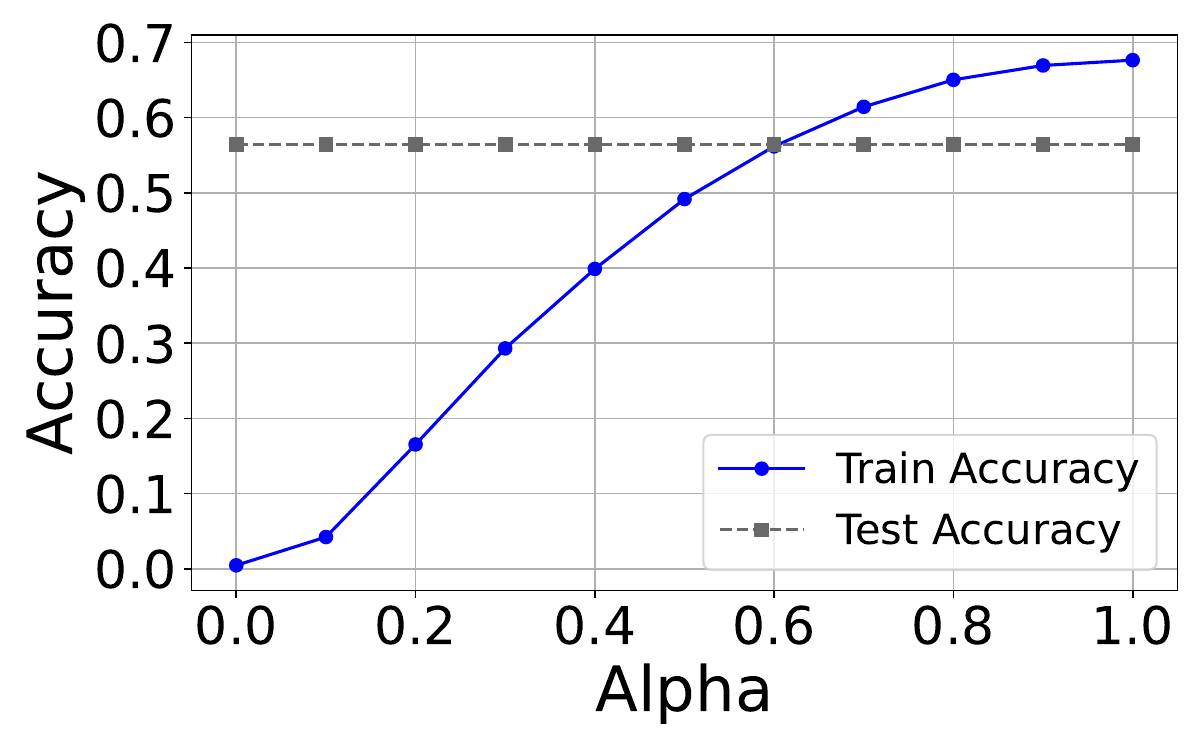}
        \vspace{-20pt}
        \caption{Accuracy vs $\alpha$}
        \label{fig:line2}
    \end{minipage}
\end{figure}

\begin{figure}[htbp]
    \centering
    \begin{minipage}[t]{0.48\linewidth}
        \centering
        \includegraphics[width=\linewidth]{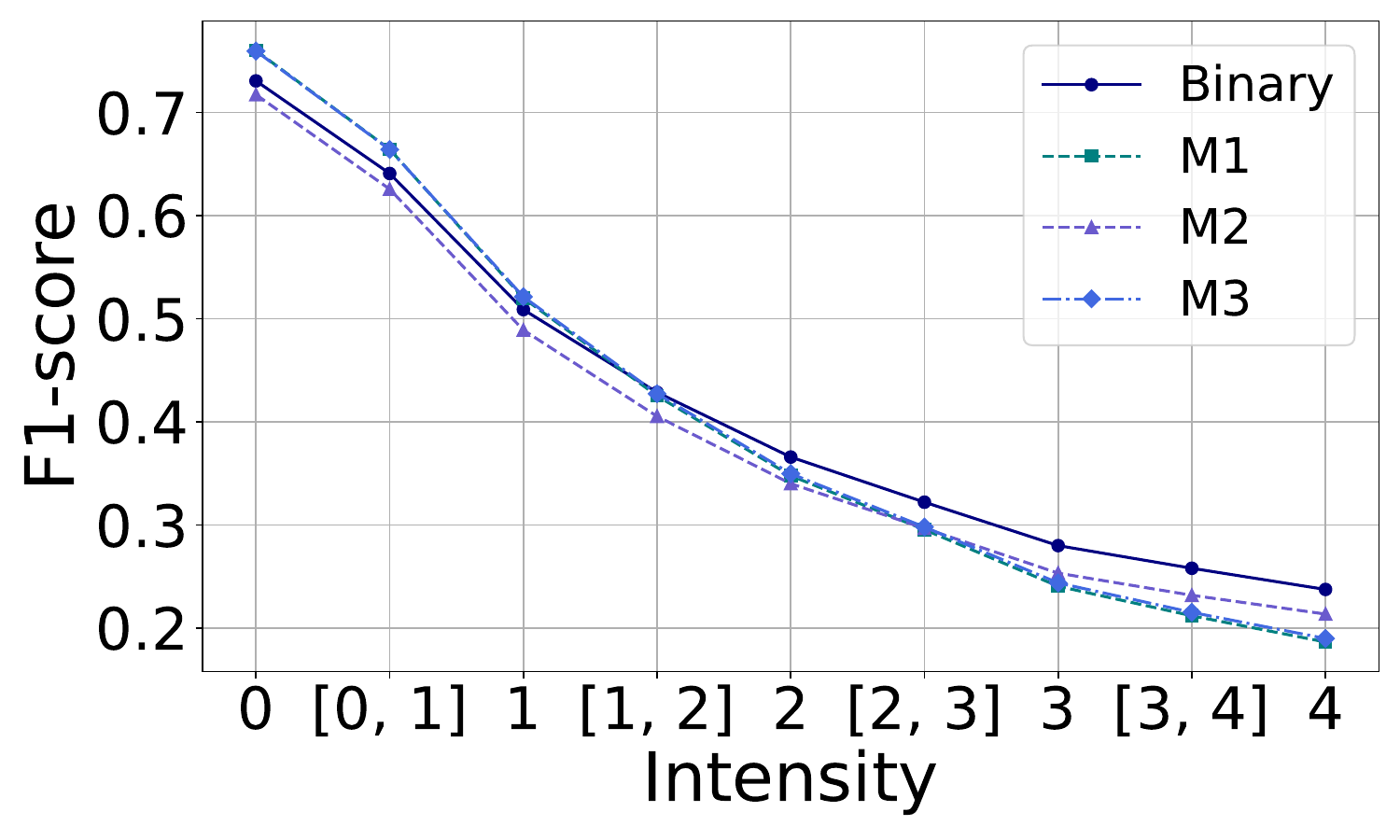}
        \vspace{-20pt}
        \caption{F1-score vs Intensity}
        \label{fig:line3}
    \end{minipage}
    \hfill
    \begin{minipage}[t]{0.48\linewidth}
        \centering
        \includegraphics[width=\linewidth]{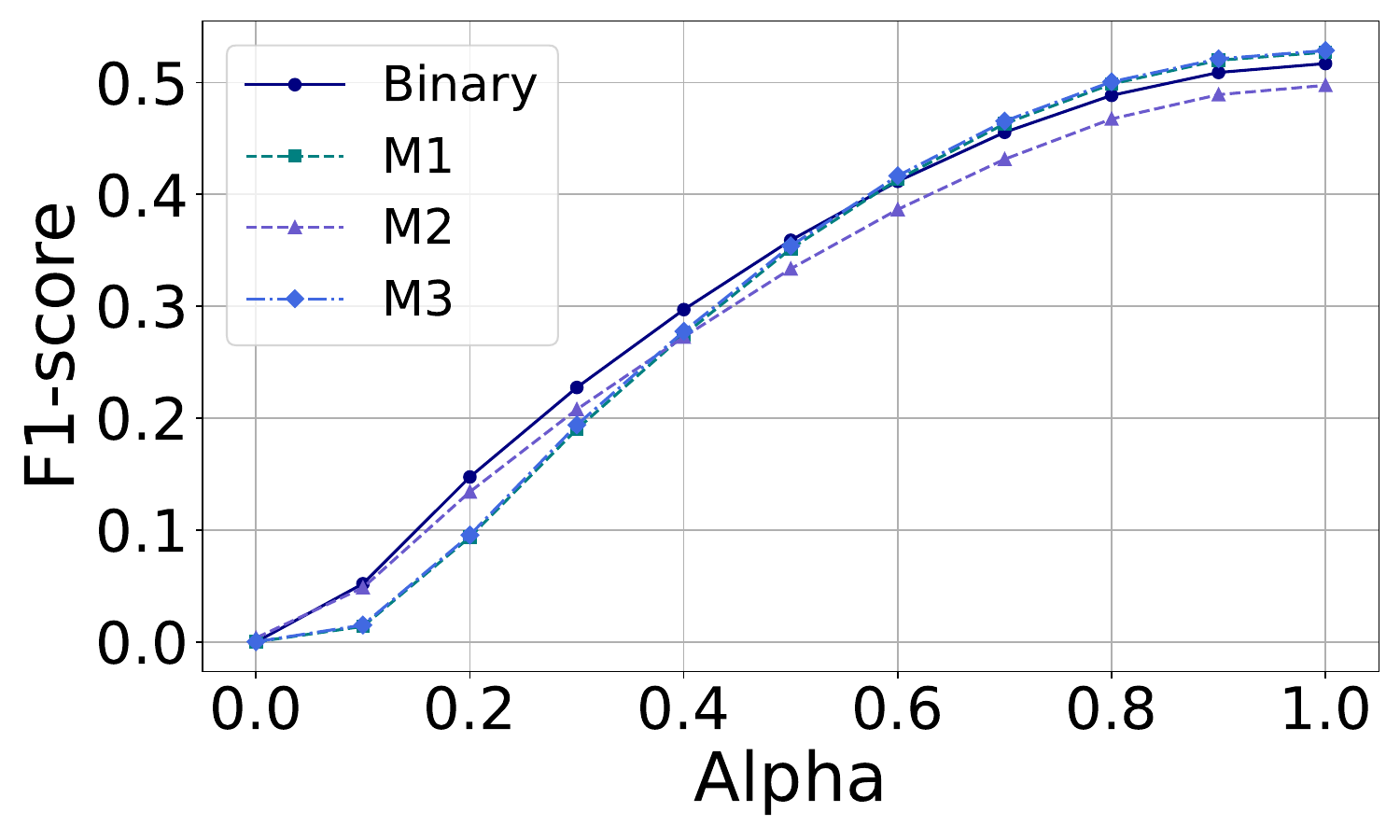}
        \vspace{-20pt}
        \caption{F1-score vs $\alpha$}
        \label{fig:line4}
    \end{minipage}
\end{figure}

Taking the Tiny-ImageNet dataset under a fully connected topology as an example, Figure~\ref{fig:line1} shows that the prediction accuracy on training samples decreases as the data augmentation intensity increases, while Figure~\ref{fig:line2} shows that it increases with higher values of $\alpha$. In both figures, the test accuracy remains unchanged, indicating that \textit{AugMixCloak} does not affect the prediction performance on benign users. Figure~\ref{fig:line3} illustrates that the F1-scores of both the binary classifier-based MIA and the metric-based MIAs decrease with increasing data augmentation intensity, whereas Figure~\ref{fig:line4} shows that these F1-scores increase as $\alpha$ increases. Note that in Figures~\ref{fig:line1} and~\ref{fig:line2}, $\alpha$ is fixed, while in Figures~\ref{fig:line3} and~\ref{fig:line4}, the data augmentation intensity is fixed. The fixed values correspond to the defense parameters shown in Table~\ref{tab:defense_params}.

Therefore, by adjusting the values of data augmentation intensity and $\alpha$, all the four F1-scores can be brought as close to 0.5 as possible.

All these findings holds consistently across all datasets and different topologies.

\subsection{Comparison with Regularization}

This paper compares \textit{AugMixCloak} with confidence score masking to better understand the defense effectiveness of \textit{AugMixCloak}. Specifically, L2 regularization is employed.

By tuning the value of \textit{weight\_decay} over a range of $0$, $1 \times 10^{-4}$, $5 \times 10^{-4}$, $1 \times 10^{-3}$, $5 \times 10^{-3}$, and $1 \times 10^{-2}$, and then refining the search within smaller intervals around promising candidates, this paper identifies the most effective weight decay values for defense as 0.003 for CIFAR-10 and 0.001 for Fashion-MNIST.

The L2 regularization is only applied to CIFAR-10 and Fashion-MNIST, which use lightweight CNN models for classification task. Because CIFAR-100, Tiny-ImageNet, and ImageNet-10 use VGG16, ResNet50, and ResNet18 respectively, which already incorporate regularization mechanisms. 

\begin{figure}[htbp]
    \centering
    \begin{minipage}[t]{0.48\linewidth}
        \centering
        \includegraphics[width=\linewidth]{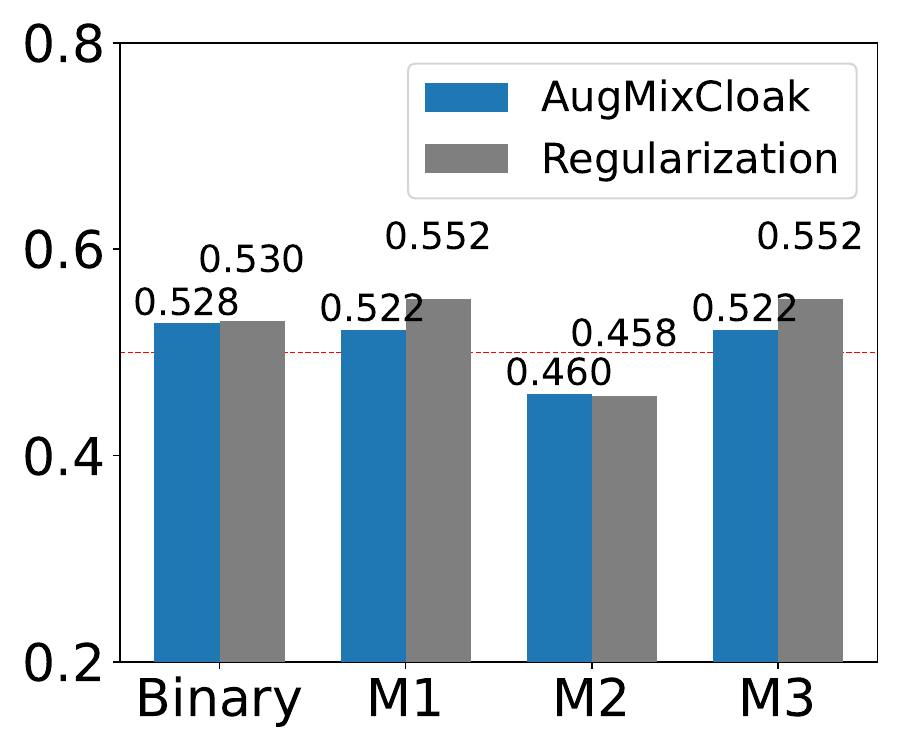}
        \vspace{-20pt}
        \caption{Comparison in CIFAR10}
        \label{fig:comp1}
    \end{minipage}
    \hfill
    \begin{minipage}[t]{0.48\linewidth}
        \centering
        \includegraphics[width=\linewidth]{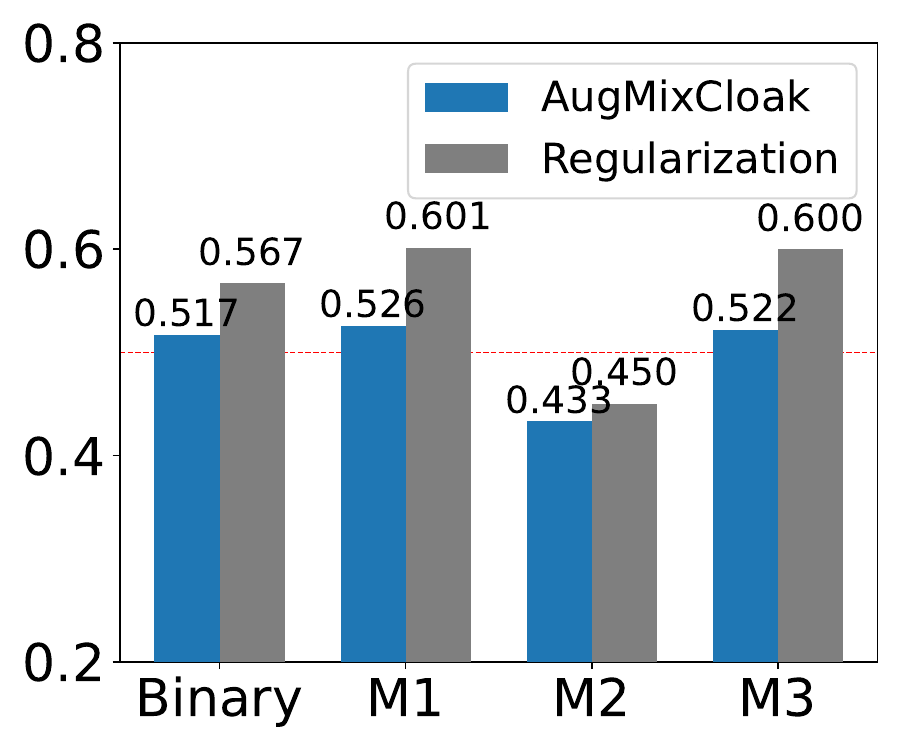}
        \vspace{-20pt}
        \caption{Comparision in Fashion-MNIST}
        \label{fig:comp2}
    \end{minipage}
\end{figure}

The comparison results are shown in Figures~\ref{fig:comp1} and~\ref{fig:comp2}. As illustrated in Figure~\ref{fig:comp1}, the four blue bars are visually closer to the red line than the adjacent gray bars. This indicates that the F1-scores of the binary classifier-based MIA and the three metric-based MIAs under \textit{AugMixCloak} are closer to 0.5, suggesting that \textit{AugMixCloak} has stronger defense performance than regularization. Regularization also demonstrates a certain level of defense effectiveness, as the values of the four metrics mostly fall within the range of 0.4 to 0.6. 

A similar finding can also be observed in Figure~\ref{fig:comp2}, except that the blue bar corresponding to M2 deviates more from 0.5 compared to the gray bar. This is because the script searches for defense parameters from a global perspective, which may cause the deviation of a single metric to be larger than that under regularization. However, if the optimization were targeted solely at the M2 metric, it would be easy to bring it close to 0.5, as shown in Figures~\ref{fig:line3} and~\ref{fig:line4}.

\subsection{Comparison with Confidence Score Masking}

This paper also compares \textit{AugMixCloak} with confidence score masking. Specifically, this paper adopts a clipping strategy that processes the model's prediction vectors by limiting the maximum predicted probability to a predefined value \textit{max\_conf}, and the remaining probabilities are redistributed proportionally.

For all four datasets including CIFAR-10, CIFAR-100, Fashion-MNIST and ImageNet-10, tuning the value of \textit{max\_conf} quickly leads to a configuration in which the F1-scores of both the binary classifier-based MIA and the three metric-based MIAs drop to zero or converge to 0.5. This suggests that both types of attacks either classify all samples as non-members or perform no better than random guessing, indicating that they are effectively mitigated.

For Tiny-ImageNet, it is possible to find a \textit{max\_conf} value that successfully defend against all three metric-based MIAs, but not the binary classifier-based MIA. This is because Tiny-ImageNet has 200 classes and contains images more visually complex than CIFAR-10, CIFAR-100, and Fashion-MNIST, providing the binary classifier-based MIA with more exploitable patterns.

Therefore, \textit{AugMixCloak} demonstrates a significant advantage, as its stronger generalization capability enables it to defend against both types of MIA across a wider range of datasets.

\section{Conclusion and Future Work}


This paper proposes and evaluates \textit{AugMixCloak}, a novel approach that mitigates MIAs by dynamically transforming query images identified as suspicious via pHash. Instead of modifying the training process or model architecture, this defense operates during the test phase by applying data augmentation and PCA-based information fusion to query images. Extensive experiments across five datasets and various DFL topologies show that \textit{AugMixCloak} is effective. It outperforms regularization in terms of effectiveness and surpasses confidence score masking in terms of generalization.

Future work will focus on designing a more refined defense approach to provide even greater protection against MIAs. In addition, the integration of \textit{AugMixCloak} with existing defense approaches will be explored to achieve better defense effectiveness.






\bibliography{mybibfile}



\end{document}